\documentclass[letterpaper]{article} 
\usepackage{aaai2027}  
\usepackage[hyphens]{url}  
\usepackage{graphicx} 
\usepackage{natbib}  
\usepackage{caption} 
\usepackage{algorithm}
\usepackage{algorithmic}
\usepackage{multirow}
\usepackage{amssymb}
\usepackage{newfloat}
\usepackage{listings}
\DeclareCaptionStyle{ruled}{labelfont=normalfont,labelsep=colon,strut=off} 
\floatstyle{ruled}
\newfloat{listing}{tb}{lst}{}
\floatname{listing}{Listing}

\usepackage{booktabs}

\title{BtsPruner: Role-Aware Visual Token Pruning for Large Vision-Language Models}
\author{
    Written by AAAI Press Staff\textsuperscript{\rm 1}\thanks{With help from the AAAI Publications Committee.}\\
    AAAI Style Contributions by Peter Patel Schneider,
    Sunil Issar,\\
    J. Scott Penberthy,
    George Ferguson,
    Hans Guesgen,
    Francisco Cruz\equalcontrib\corresponding,
    Marc Pujol-Gonzalez\equalcontrib\corresponding
}
\affiliations{
    \textsuperscript{\rm 1}Association for the Advancement of Artificial Intelligence\\

    1101 Pennsylvania Ave, NW Suite 300\\
    Washington, DC 20004 USA\\
    proceedings-questions@aaai.org
}

\title{BtsPruner: Role-Aware Visual Token Pruning for Large Vision-Language Models}
\author {
    Zihao Qi\textsuperscript{\rm 1,\rm 2}\equalcontrib,
    Zhibo Zhang\textsuperscript{\rm 1,\rm 2}\equalcontrib,
    Miaoxin Zhang\textsuperscript{\rm 2}\equalcontrib,
    Jin Cheng Cai\textsuperscript{\rm 2}\equalcontrib,
    Zengqiang Yan\textsuperscript{\rm 1,\rm 2}\corresponding
}
\affiliations {
    \textsuperscript{\rm 1}Nanjing University\\
    \textsuperscript{\rm 2}Huazhong University of Science and Technology\\
    qizihao@smail.nju.edu.cn, z\_yan@hust.edu.cn
}

\begin{document}

\maketitle

\begin{abstract}
    Large Vision-Language Models (LVLMs) often process hundreds or thousands of visual tokens, resulting in substantial computational and memory overhead during inference. Existing training-free pruning methods typically estimate token importance using language-model attention, visual attention, or feature similarity. However, these methods treat visual tokens with uniform criteria and overlook that tokens may serve different roles during visual encoding. 
    In this work, we investigate the evolution of visual token roles in LVLMs and observe that tokens progressively transition from local detail carriers to globally referenced representatives as encoding proceeds. Based on this observation, we propose \textbf{BtsPruner}, a training-free role-aware visual token pruning framework that preserves complementary token roles under a fixed budget. Specifically, BtsPruner first identifies the token-role transition point based on the evolution of attention concentration. It then discovers globally referenced representatives from later encoding stages and recursively retrieves local detail tokens from earlier stages to recover complementary fine-grained evidence.
    Extensive experiments across LLaVA-1.5, LLaVA-NeXT, Qwen-VL, InternVL, and CogVLM demonstrate that BtsPruner achieves superior performance retention under aggressive token reduction. On LLaVA-1.5-7B, BtsPruner preserves \textbf{91.6\%} of the dense model's performance using only \textbf{32 of 576 visual tokens (5.6\%)}. In the high-resolution LLaVA-NeXT-7B setting, it achieves up to a \textbf{5.95$\times$ CUDA-latency speedup}.
\end{abstract}

\begin{links}
    \link{Code}{https://github.com/Halenbest/BtsPruner}
\end{links}

\begin{figure}[!t]
    \centering
    \includegraphics[width=\linewidth]{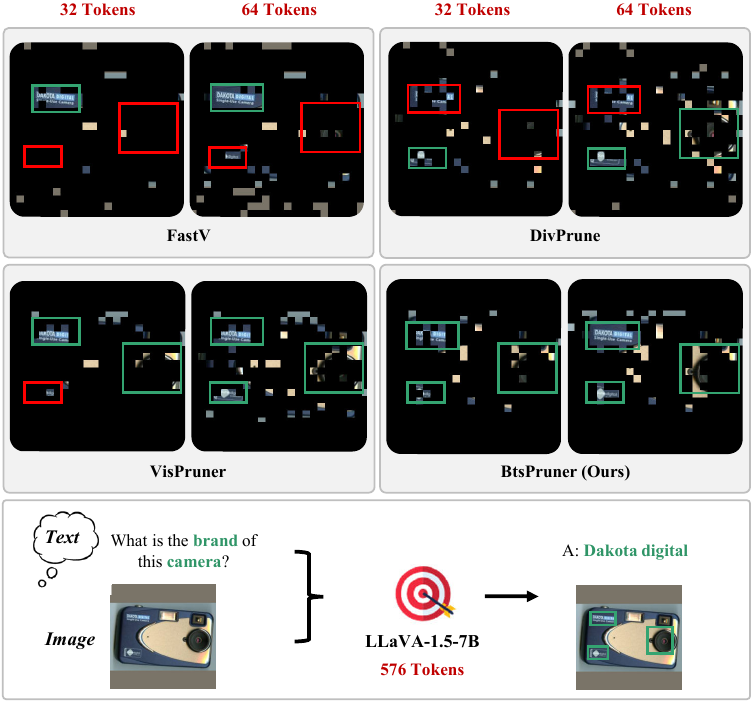}
    \caption{
        Qualitative comparison of visual token pruning strategies. Existing methods select tokens according to a unified criterion, while BtsPruner preserves complementary token roles under the same token budget.
    }
    \label{fig1}
\end{figure}

\section{Introduction}

Large vision-language models (LVLMs) extend large language models (LLMs) to multimodal understanding and reasoning
\cite{bai2023qwen,jiang2023mistral, touvron2023llamaopenefficientfoundation,bai2023qwenvl, Cao_2025_CVPR, 10656429}.
A typical LVLM encodes an image into visual tokens, projects them into the language-model embedding space, and jointly processes them with textual tokens. However, visual tokens often dominate the input sequence, substantially increasing computation, key-value cache usage, and inference latency, especially in high-resolution scenarios.

Training-free visual token pruning provides a practical solution without updating model parameters. FastV~\cite{chen2024fastv} and subsequent methods \cite{xing2024pyramiddrop, ye2024fitprune, zhang2024sparsevlm} use text-to-visual attention in the language model, but such attention may shift or disperse across layers \cite{xu2026rethinkingvisualtokenreduction}. Recent methods therefore perform pruning near the vision encoder using visual attention, feature similarity, or their combination \cite{shang2024llavaprumerge,yang2024visionzip, 11095009,11444044}. Nevertheless, most existing methods \textit{estimate token importance with a unified criterion} and overlook that visual tokens may develop different roles during encoding, leading to the loss of either globally representative information or fine-grained local details under aggressive pruning as illustrated in Fig.~\ref{fig1}.

To investigate this issue, we analyze the evolution of visual token roles during LVLM encoding. We observe that visual tokens gradually transition from local detail carriers to globally referenced representatives. In early encoding stages, tokens mainly interact with nearby or visually related regions and preserve localized visual details. As encoding progresses, a subset of tokens becomes increasingly shared across visual queries and provides broader contextual representations. This role evolution suggests that \textit{effective pruning should preserve complementary token roles rather than select tokens solely according to a single importance score}.

Based on this observation, we propose \textbf{BtsPruner}, a training-free role-aware visual token pruning framework. BtsPruner first identifies the token role transition according to attention concentration changes. It then discovers globally referenced representatives through attention-guided selection and diversity filtering, and recursively retrieves local detail tokens to preserve complementary fine-grained evidence. By maintaining both token roles under a fixed budget, BtsPruner reduces redundant visual tokens while preserving visual coverage. We evaluate BtsPruner on LLaVA-1.5~\cite{liu2024llava1.5}, LLaVA-NeXT~\cite{liu2024llavanext}, Qwen-VL~\cite{bai2023qwenvl}, InternVL~\cite{10656429}, and CogVLM~\cite{wang2025cogvlm}. Extensive experiments across multimodal benchmarks demonstrate that BtsPruner achieves superior performance retention under aggressive token compression. On LLaVA-1.5-7B, BtsPruner preserves \textbf{91.6\%} of the dense model's performance using only \textbf{32 of 576 visual tokens (5.6\%)}. In high-resolution LLaVA-NeXT-7B settings, it achieves up to a \textbf{5.95$\times$ CUDA-latency speedup}.
The main contributions are summarized as follows:
\begin{itemize}
    \item We reveal the role evolution phenomenon of visual tokens in LVLM encoders, where tokens transition from local detail carriers to globally referenced representatives during encoding.
    \item We propose BtsPruner, a role-aware visual token pruning framework that discovers and preserves complementary token roles under a fixed token budget.
    \item Extensive experiments across multiple LVLM architectures and benchmarks demonstrate strong performance retention, particularly under extreme token compression.
\end{itemize}

\begin{figure*}[t]
    \centering
    \includegraphics[width=\textwidth]{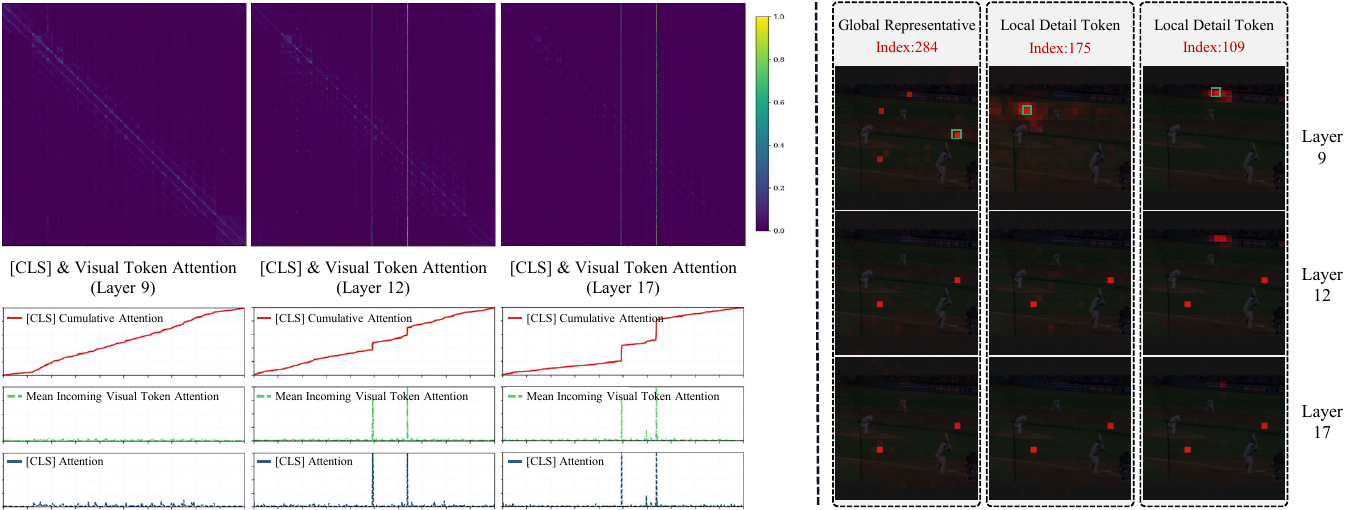} 
    \caption{
        Visual token role evolution during LVLM encoding. \textit{Left}: Attention patterns show a transition from localized interactions to shared representations. \textit{Right}: Representative local detail and global representative tokens across different encoding stages.
    }
    \label{fig2}
\end{figure*}

\section{Related Work}

\subsection{Large Vision-Language Models}

Large vision-language models (LVLMs) extend large language models to multimodal understanding and generation tasks
\cite{bai2023qwen, ouyang2022instructgpt, achiam2023gpt, liu2024llava, team2023gemini, wang2025cogvlm}. Most LVLMs adopt a modular architecture consisting of a vision encoder, a multimodal projector, and a large language model. The vision encoder converts images into visual token sequences, while the projector aligns visual representations with the language-model embedding space. Common projector designs include linear projection~\cite{liu2024llava}, MLP-based mapping~\cite{liu2024llava1.5}, and query-based modules~\cite{bai2023qwenvl, li2023blip2}. Although increasing image resolution improves multimodal perception ability~\cite{liu2024llavanext}, it also introduces substantially longer visual-token sequences, increasing computational and memory costs. Therefore, efficient processing of visual tokens has become an important problem for LVLM deployment.

\subsection{Visual Token Pruning in LVLMs}

Visual token pruning aims to reduce redundant visual representations while preserving task-relevant information during LVLM inference. Existing approaches can be broadly categorized into training-based and training-free methods. Training-based methods introduce additional token compression or resampling modules \cite{cai2024m3, chen2024llavolta, li2024tokenpacker}, but require extra training costs and modifications to existing models. In contrast, training-free methods directly estimate token importance and can be applied to pretrained LVLMs.

Early training-free methods mainly perform pruning within the language model. FastV~\cite{chen2024fastv} introduces text-to-visual attention-based pruning, followed by several approaches that adopt similar strategies \cite{xing2024pyramiddrop, ye2024fitprune, zhang2024sparsevlm}. However, language-model-side attention may suffer from attention shift and dispersion across layers \cite{xu2026rethinkingvisualtokenreduction}. Recent studies therefore move pruning closer to the vision encoder and exploit visual attention, feature similarity, or their combination. DivPrune~\cite{11095009} selects diverse tokens based on feature similarity, while PruMerge, VisionZip, and VisPruner \cite{shang2024llavaprumerge, yang2024visionzip, 11444044} combine attention-based importance estimation with similarity-based redundancy reduction.

Despite their effectiveness, existing training-free methods generally assume that visual tokens can be ranked using a unified importance criterion. Such strategies overlook that visual tokens may gradually develop different roles during encoding, where some tokens preserve local details while others become globally referenced representatives. This motivates role-aware visual token pruning that preserves complementary token roles rather than relying on a single importance estimation criterion.

\section{Visual Token Role Evolution in LVLMs}

Existing visual token pruning methods usually estimate token importance using a unified criterion, assuming that visual tokens are interchangeable during encoding. However, visual tokens may gradually develop different functional roles as the representation evolves. In this section, we investigate the evolution of visual token roles during LVLM encoding and identify two complementary roles: \textbf{local detail tokens}, which preserve fine-grained visual evidence, and \textbf{global representative tokens}, which provide broader contextual coverage. We first introduce the self-attention mechanism and the statistics used for analysis, then analyze the role transition of visual tokens and discuss its implications for role-aware token pruning.

\subsection{Preliminaries}

Most vision encoders used in LVLMs, such as CLIP~\cite{radford2021clip}, are based on the Transformer architecture. Given an input token sequence
\begin{equation}
    X=[x_{\mathrm{cls}};x_v^1,\ldots,x_v^n]\in\mathbb{R}^{(n+1)\times d},
\end{equation}
where $x_{\mathrm{cls}}$ denotes the [CLS] token, $x_v^i$ denotes the $i$-th visual token, $n$ is the number of visual tokens, and $d$ is the hidden dimension, the query, key, and value representations are computed as
\begin{equation}
    Q=XW_Q,\qquad K=XW_K,\qquad V=XW_V,
\end{equation}
where $W_Q,W_K,W_V\in\mathbb{R}^{d\times d}$ are learnable projection matrices. The scaled dot-product attention and its output are calculated as
\begin{equation}
    \label{OAV}
    A=\mathrm{softmax}\left(\frac{QK^\top}{\sqrt{d}}\right),
    \qquad O=AV.
\end{equation}
We refer to the first row of $A$ as the [CLS] attention. By removing the [CLS] row and column, we obtain the visual-token attention matrix $A'\in\mathbb{R}^{n\times n}$. The $i$-th row, $A'_{i,:}$, represents the attention distribution of the $i$-th visual token over other visual tokens. To characterize whether a token gradually becomes a shared representation, we define the mean incoming attention of token $j$ as
\begin{equation}
    \bar{a}_j=\frac{1}{n}\sum_{i=1}^{n}A'_{ij},
\end{equation}
where $\bar{a}_j$ measures the degree to which token $j$ is referenced by other visual-token queries. A higher value indicates that the token receives attention from more visual tokens and may serve as a shared representation. For multi-head self-attention, we average the attention matrices across all heads before analysis.

\subsection{Token Role Transition Analysis}

To investigate whether visual tokens develop different roles during encoding, we extract attention scores from the vision encoder of LLaVA-1.5-7B~\cite{liu2024llava1.5} using a representative example from VQAv2~\cite{goyal2017vqav2}. Figure~\ref{fig2} presents the evolution of visual-token attention across different encoder layers. As illustrated in Fig.~\ref{fig2}, in early encoding stages (\textit{e.g.}, Layer 9), the attention matrix exhibits localized diagonal structures, indicating that visual tokens mainly interact with nearby or visually related regions. We refer to these tokens as \textbf{local detail tokens} since they preserve spatially specific visual patterns. As encoding progresses, attention gradually becomes shared across queries and concentrates on a smaller set of commonly referenced positions in deeper layers (\textit{e.g.}, Layer 17). The attention statistics further support this transition. The [CLS] cumulative-attention curve changes from a uniform distribution to a concentrated distribution, and [CLS] attention exhibits similar peaks with mean incoming visual-token attention. 
Visualization in Fig.~\ref{fig2} further shows that token 284 becomes a  \textbf{global representative token}, while tokens 175 and 109 preserve local details. These observations indicate that visual tokens evolve from local detail carriers to globally referenced representatives during encoding.

\subsection{Implications for Role-Aware Token Pruning}

Attention weights do not provide a complete causal explanation of representation formation, as token representations are also affected by value projections, residual connections, and feed-forward networks. Nevertheless, self-attention reflects token interactions within Transformer blocks.
The observed role evolution suggests that visual token pruning should preserve complementary roles rather than rank all tokens using a single criterion. Selecting only global representative tokens may lose fine-grained details, while selecting only local detail tokens may introduce redundancy and insufficient global context.
Therefore, effective pruning should preserve both token roles under a limited budget. This motivates BtsPruner to discover global representative tokens and retrieve complementary local detail tokens for role-aware visual token compression.

\begin{figure*}[t]
    \centering
    \includegraphics[width=\textwidth]{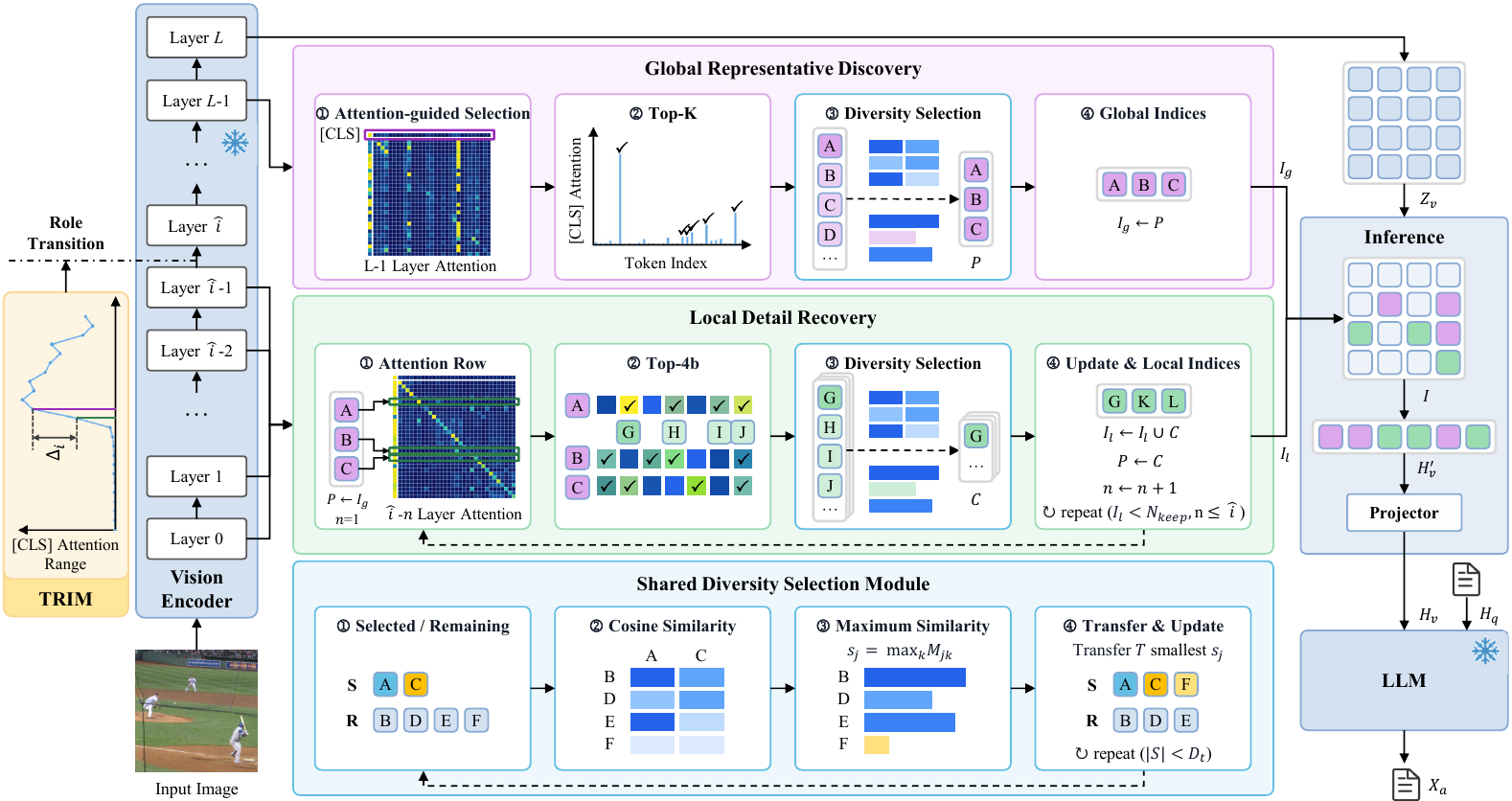} 
    \caption{
        Overview of BtsPruner. BtsPruner first identifies the token role transition during visual encoding, then discovers global representative tokens and recursively recovers local detail tokens through a shared diversity selection module. The selected tokens with complementary roles are combined for efficient LVLM inference.
    }
    \label{fig4}
\end{figure*}

\section{Method}

In this section, we introduce BtsPruner, a training-free role-aware visual token pruning framework. As illustrated in Fig.~\ref{fig4}, BtsPruner first identifies the transition of token roles during encoding. It then discovers global representative tokens through attention-based selection and diversity filtering, and recursively retrieves local detail tokens to preserve complementary visual evidence. Finally, tokens with different roles are combined for inference.

\subsection{Role Transition Identification}

Visual tokens gradually evolve from local detail carriers to global representatives during encoding. Therefore, BtsPruner first identifies the transition point between these roles through a token role transition identification module (TRIM). As shown in Fig.~\ref{fig2}, the [CLS] cumulative attention is relatively uniform in earlier layers but becomes concentrated on several tokens in later layers. We use this change in attention concentration to identify the transition point.
At layer $i$, the [CLS] attention is denoted as $\mathrm{CLS}_i\in\mathbb{R}^{n}$.
We calculate its range as:
\begin{equation}
    \mathrm{range}_i = \max(\mathrm{CLS}_i)-\min(\mathrm{CLS}_i)
\end{equation}
The change between adjacent layers is defined as:
\begin{equation}
    \Delta_i = \mathrm{range}_i - \mathrm{range}_{i-1}
\end{equation}
Then, the transition point is determined by:
\begin{equation}
\widehat{i}=\{i|\Delta_i\geq(\mathrm{CLS}_{max}-\mathrm{CLS}_{min})/U\},
\end{equation}
where $\mathrm{CLS}_{max}$ and $\mathrm{CLS}_{min}$ denote the maximum and minimum values of [CLS] attention across all encoder layers, respectively. $U$ controls the threshold and is set to 2 by default.
The layers before $\widehat{i}$ are used for local detail recovery, while later layers are used for global representative discovery.
For vision encoders without a [CLS] token, the mean incoming attention of visual tokens is used instead.

\subsection{Global Representative Discovery}

As encoding progresses, some visual tokens become commonly referenced by multiple visual queries and provide global contextual representations. BtsPruner discovers these global representatives using attention-guided selection and diversity filtering.
Given pruning ratio $R$, we first select candidate tokens according to the attention of the second-to-last encoding layer. Given $Z_v$ as the visual token features output by the vision encoder, we select candidate tokens as:
\begin{equation}
    Z_{\mathrm{global}}^{cand} =\mathrm{TopK}(Z_v,\mathrm{CLS}_{-2}),
\end{equation}
where $\mathrm{CLS}_{-2}$ denotes the [CLS] attention from the second-to-last layer, and $K=n(1-R)$ is the number of selected candidate tokens. Following previous studies~\cite{yang2024visionzip,11444044}, we use the second-to-last layer as the default representative discovery layer. The candidate tokens are further filtered by the diversity selection module:
\begin{equation}
    P=\mathrm{Diverse}(Z_{\mathrm{global}}^{cand},N_{\mathrm{global}},T),
\end{equation}
where $P$ denotes the selected global representative token set, and $T$ denotes the number of candidate tokens transferred in each iteration of the diversity selection process.
$N_{\mathrm{global}}$ is defined as $n(1-R)r$ where $n(1-R)$ represents the total number of retained tokens and $r$ determines the proportion allocated to global representatives. $N_{\mathrm{global}}$ is also the target size $D_t$ of the diversity selection module. The selected indices form the global representative set $I_g$.

\begin{table*}[!t]
    \centering
    \small
    \setlength{\tabcolsep}{1.0mm}
    \resizebox{0.9\textwidth}{!}{
        \begin{tabular}{@{}l|cccccccccc|cc@{}}
            \toprule
            Method & VQAv2 & GQA & VizWiz & SQA$^{\mathrm{IMG}}$ & VQA$^{\mathrm{Text}}$ & POPE & MME & MMB & MMB$^{\mathrm{CN}}$ & MMVet & Avg. & Rel. (\%) \\
            \midrule
            \multicolumn{13}{c}{Upper Bound (576 tokens)} \\
            \cmidrule{1-13}
            LLaVA-1.5-7B & 78.5 & 62.0 & 50.0 & 66.8 & 58.2 & 85.9 & 1510.7 & 64.3 & 58.3 & 31.1 & 63.1 & 100.0\% \\
            \midrule
            \multicolumn{13}{c}{Retain 128 Tokens ($\downarrow$ 77.8\%)} \\
            \cmidrule{1-13}
            FastV~(ECCV24) & 73.2 & 55.8 & 51.5 & 68.2 & 56.9 & 68.8 & 1442.3 & 63.2 & 56.7 & 28.0 & 59.4 & 94.7\% \\
            SparseVLM~(ICML25) & 76.2 & 58.4 & 50.0 & 65.2 & 56.8 & 85.2 & 1380.4 & 63.7 & 58.1 & 28.3 & 61.1 & 96.7\% \\
            DivPrune~(CVPR25) & 75.7 & 59.3 & 53.3 & 68.0 & 55.5 & 84.3 & 1407.7 & 61.1 & 54.4 & 28.7 & 61.1 & 96.8\% \\
            PruMerge+~(2024.05) & 74.9 & 57.9 & 52.9 & 65.2 & 54.8 & 84.2 & 1402.5 & 60.2 & 55.6 & 30.0 & 60.6 & 96.3\% \\
            VisionZip~(CVPR25) & 75.3 & 57.5 & 52.6 & 68.4 & 56.3 & 83.0 & 1434.4 & 62.3 & 56.2 & 29.8 & 61.3 & 97.4\% \\
            VisPruner~(ICCV25) & 75.7 & 58.2 & 52.6 & 69.0 & 57.2 & 84.3 & 1430.7 & 60.9 & 55.4 & 30.6 & 61.5 & 97.8\% \\
            BtsPruner~(Ours) & 76.1 & 58.5 & 52.7 & 69.0 & 57.7 & 84.8 & 1439.2 & 61.4 & 56.3 & 31.6 & \textbf{62.0} & \textbf{98.7\%} \\
            \midrule
            \multicolumn{13}{c}{Retain 64 Tokens ($\downarrow$ 88.9\%)} \\
            \cmidrule{1-13}
            FastV~(ECCV24) & 66.3 & 51.5 & 51.6 & 68.0 & 55.0 & 64.1 & 1235.1 & 60.2 & 52.7 & 28.2 & 55.9 & 89.8\% \\
            SparseVLM~(ICML25) & 70.2 & 53.8 & 50.1 & 66.3 & 53.4 & 78.4 & 1252.3 & 59.4 & 52.5 & 24.1 & 57.1 & 90.2\% \\
            DivPrune~(CVPR25) & 73.9 & 57.4 & 53.8 & 67.9 & 54.9 & 81.0 & 1357.6 & 59.2 & 51.6 & 27.7 & 59.5 & 94.4\% \\
            PruMerge+~(2024.05) & 71.8 & 55.5 & 53.4 & 65.0 & 54.0 & 80.1 & 1348.5 & 58.2 & 53.0 & 26.7 & 58.5 & 92.8\% \\
            VisionZip~(CVPR25) & 72.1 & 55.0 & 53.1 & 68.8 & 55.2 & 77.0 & 1353.1 & 59.4 & 54.6 & 29.7 & 59.3 & 94.5\% \\
            VisPruner~(ICCV25) & 72.7 & 56.0 & 53.4 & 68.5 & 55.8 & 80.1 & 1365.9 & 58.6 & 54.3 & 30.1 & 59.8 & 95.3\% \\
            BtsPruner~(Ours) & 73.4 & 56.3 & 53.4 & 69.6 & 56.3 & 80.2 & 1364.4 & 60.1 & 54.6 & 31.4 & \textbf{60.4} & \textbf{96.4\%} \\
            \midrule
            \multicolumn{13}{c}{Retain 32 Tokens ($\downarrow$ 94.4\%)} \\
            \cmidrule{1-13}
            FastV~(ECCV24) & 56.9 & 46.7 & 49.2 & 66.2 & 51.9 & 57.3 & 995.5 & 51.2 & 43.6 & 20.6 & 49.3 & 78.8\% \\
            SparseVLM~(ICML25) & 64.5 & 50.2 & 48.1 & 66.9 & 49.9 & 70.7 & 1130.9 & 56.8 & 46.3 & 20.7 & 53.1 & 83.7\% \\
            DivPrune~(CVPR25) & 70.1 & 53.9 & 52.9 & 67.3 & 52.4 & 78.5 & 1301.4 & 56.9 & 49.2 & 26.2 & 57.2 & 90.7\% \\
            PruMerge+~(2024.05) & 66.6 & 51.7 & 52.8 & 65.6 & 52.2 & 74.6 & 1203.1 & 55.6 & 48.5 & 25.5 & 55.3 & 88.0\% \\
            VisionZip~(CVPR25) & 66.8 & 51.6 & 52.4 & 68.7 & 52.7 & 68.3 & 1229.3 & 56.2 & 50.4 & 27.3 & 55.6 & 88.9\% \\
            VisPruner~(ICCV25) & 67.6 & 52.0 & 52.5 & 69.1 & 53.5 & 72.9 & 1258.6 & 56.8 & 48.5 & 25.1 & 56.1 & 89.1\% \\
            BtsPruner~(Ours) & 69.0 & 53.6 & 53.1 & 69.5 & 54.5 & 74.3 & 1292.2 & 58.0 & 51.6 & 27.1 & \textbf{57.5} & \textbf{91.6\%} \\
            \bottomrule
        \end{tabular}
    }
    \caption{
        Comparison of visual token pruning methods on LLaVA-1.5-7B across three pruning rates. Avg. denotes the normalized average score across benchmarks, and Rel. denotes the relative performance retention ratio compared with the dense model.
    }
    \label{tab:baseline}
\end{table*}

\subsection{Diversity Selection Module}

The diversity selection module is shared by global representative discovery and local detail recovery. Given candidate features, we first initialize a selected set $S$ and remaining set $\mathcal{R}$. The cosine similarity between selected and remaining features is calculated, and candidates with the lowest maximum similarity to $S$ are iteratively transferred into $S$. This process continues until the selected token number reaches the predefined target size $D_t$.

\subsection{Local Detail Recovery}

Although global representatives provide broad contextual coverage, they may omit fine-grained visual details. Therefore, BtsPruner further retrieves complementary local detail tokens. Starting from layer $\widehat{i}-1$, the selected global representatives $I_g$ are treated as parent tokens. For each parent token, we select the top-$4b$ candidates according to its attention distribution:
\begin{equation}
    Z_{\mathrm{local}}^{cand}=\mathrm{Top}_{4b}(Z_v,A_p),
\end{equation}
where $A_p$ denotes the attention row of its parent token. The candidate tokens are then filtered using the diversity selection module:
\begin{equation}
    C=\mathrm{Diverse}(Z_{\mathrm{local}}^{cand},b,T),
\end{equation}
where $b$ denotes the number of local detail tokens assigned to each parent token, and $C$ represents the local detail tokens selected from the candidates of the current parent token. The selected local detail tokens are added to $I_l$ and become parent tokens for the next recovery step. This recursive process continues until the target token budget $N_{keep}$ is reached or the earliest layer is reached.

\subsection{Inference with Role-Aware Token Pruning}

The final selected indices are obtained by combining global and local tokens:
\begin{equation}
    I=\mathrm{sort}(\mathrm{unique}(I_g\cup I_l)),
    \qquad
    H_v'=Z_v[I],
\end{equation}
where $\mathrm{unique}(\cdot)$ removes duplicated token indices from the two selected sets, and $\mathrm{sort}(\cdot)$ arranges the remaining indices according to their original order in the visual token sequence.
The selected visual tokens are first mapped into the language-model embedding space through the multimodal projector $W$, and then concatenated with the text prompt $H_q$ before being fed into the language model $f_{\phi}$:
\begin{equation}
H_v=W(H_v'),
\qquad
X_a=f_{\phi}(H_v\oplus H_q),
\end{equation}
where $W$ denotes the multimodal projector that aligns visual token representations with the language-model embedding space, and $f_{\phi}$ denotes the language model parameterized by $\phi$. By preserving complementary visual token roles, BtsPruner achieves effective token compression while maintaining visual evidence for multimodal reasoning.

\section{Experiments}

\subsection{Experimental Setup}

\paragraph{Datasets.}
We evaluate BtsPruner on ten widely used multimodal benchmarks, including VQAv2~\cite{goyal2017vqav2}, GQA~\cite{hudson2019gqa}, VizWiz~\cite{gurari2018vizwiz}, ScienceQA-IMG~\cite{lu2022sqa}, TextVQA~\cite{singh2019textvqa}, POPE~\cite{li2023pope}, MME~\cite{fu2026mme}, MMBench~\cite{liu2024mmbench}, MMBench-CN~\cite{liu2024mmbench}, and MMVet~\cite{yu2023mm}. All evaluations follow the official settings and metrics.

\paragraph{Model Architectures.}
We evaluate BtsPruner on LLaVA-1.5~\cite{liu2024llava1.5}, LLaVA-NeXT~\cite{liu2024llavanext} for high-resolution scenarios, and other LVLM architectures including Qwen-VL~\cite{bai2023qwenvl}, InternVL~\cite{10656429}, and CogVLM~\cite{wang2025cogvlm}.
All models follow their original evaluation settings.

\paragraph{Comparison Methods.}
We compare with representative training-free pruning methods, including language-model-side attention methods (FastV~\cite{chen2024fastv}, SparseVLM~\cite{zhang2024sparsevlm}), similarity-based methods near the vision encoder (DivPrune~\cite{11095009}), and attention-similarity hybrid methods (PruMerge~\cite{shang2024llavaprumerge}, VisionZip~\cite{yang2024visionzip}, VisPruner~\cite{11444044}).

\paragraph{Implementation Details.}
All experiments are conducted on NVIDIA GeForce RTX 3090 GPUs. The ratio between global representatives and local detail tokens, $r$, is set to 0.5 by default. We further investigate the influence of $r$ through ablation experiments.

\begin{table}[t]
    \centering
    \footnotesize
    \setlength{\tabcolsep}{0.4mm}
    \resizebox{1\linewidth}{!}{
        \begin{tabular}{@{}l|ccccc|cc@{}}
        \toprule
        Method & VQAv2 & GQA & VQA$^{\mathrm{Text}}$ & POPE & MME & Avg. & Rel. (\%) \\
        \midrule
        \multicolumn{8}{c}{Upper Bound (2880 tokens)} \\
        \cmidrule{1-8}
        LLaVA-1.6-7B & 81.3 & 62.4 & 60.2 & 86.7 & 1511.8 & 73.2 & 100.0\% \\
        \midrule
        \multicolumn{8}{c}{Retain 640 Tokens ($\downarrow$ 77.8\%)} \\
        \cmidrule{1-8}
        FastV & 78.6 & 60.0 & 59.0 & 72.9 & 1467.6 & 68.8 & 94.4\% \\
        SparseVLM & 77.4 & 59.8 & 57.0 & 81.0 & 1457.5 & 69.6 & 95.1\% \\
        DivPrune & 77.1 & 59.8 & 54.5 & 82.4 & 1444.2 & 69.2 & 94.3\% \\
        VisionZip & 77.0 & 58.6 & 56.1 & 82.5 & 1388.8 & 68.7 & 93.8\% \\
        VisPruner & 78.8 & 61.3 & 60.1 & 85.7 & 1484.2 & 72.0 & 98.4\% \\
        Ours & 79.3 & 61.6 & 60.7 & 86.6 & 1488.5 & \textbf{72.5} & \textbf{99.1\%} \\
        \midrule
        \multicolumn{8}{c}{Retain 320 Tokens ($\downarrow$ 88.9\%)} \\
        \cmidrule{1-8}
        FastV & 73.5 & 56.0 & 56.4 & 69.0 & 1360.4 & 64.6 & 88.7\% \\
        SparseVLM & 72.1 & 54.4 & 52.8 & 73.3 & 1334.6 & 63.9 & 87.3\% \\
        DivPrune & 75.3 & 57.9 & 53.5 & 80.7 & 1393.3 & 67.4 & 91.9\% \\
        VisionZip & 74.7 & 56.7 & 55.6 & 79.5 & 1294.3 & 66.2 & 90.5\% \\
        VisPruner & 75.8 & 59.0 & 58.2 & 81.2 & 1405.8 & 68.9 & 94.2\% \\
        Ours & 76.7 & 59.5 & 59.7 & 82.3 & 1451.5 & \textbf{70.2} & \textbf{96.0\%} \\
        \midrule
        \multicolumn{8}{c}{Retain 160 Tokens ($\downarrow$ 94.4\%)} \\
        \cmidrule{1-8}
        FastV & 63.1 & 50.7 & 52.9 & 62.0 & 1123.3 & 57.0 & 78.5\% \\
        SparseVLM & 66.1 & 51.4 & 47.8 & 64.7 & 1215.3 & 58.1 & 79.6\% \\
        DivPrune & 73.2 & 56.9 & 52.5 & 78.6 & 1355.2 & 65.8 & 89.7\% \\
        VisionZip & 71.1 & 54.5 & 53.8 & 78.1 & 1144.5 & 62.9 & 86.0\% \\
        VisPruner & 71.6 & 56.0 & 56.0 & 75.3 & 1285.3 & 64.6 & 88.5\% \\
        Ours & 73.4 & 56.7 & 57.7 & 77.5 & 1358.9 & \textbf{66.6} & \textbf{91.3\%} \\
        \bottomrule
        \end{tabular}
    }
    \caption{
        Performance comparison of visual token pruning methods on LLaVA-1.6-7B, i.e., LLaVA-NeXT-7B, across three pruning rates. Avg. denotes the normalized average score across benchmarks, and Rel. denotes the relative performance retention ratio compared with the dense model.
    }
    \label{tab:highres}
\end{table}

\subsection{Main Results}

We first evaluate BtsPruner on LLaVA-1.5-7B. Table~\ref{tab:baseline} compares different pruning methods under three token budgets. BtsPruner achieves the highest Avg. and Rel. across all evaluated budgets. When retaining 128 tokens, BtsPruner preserves 98.7\% of the dense model performance. With only 64 retained tokens, it still maintains 96.4\% performance retention.
Under extreme pruning with only 32 retained tokens (5.6\% of the original visual tokens), BtsPruner preserves 91.6\% of the dense model performance. Compared with the strongest baseline VisPruner, BtsPruner improves relative performance retention by 0.9, 1.1, and 2.5 percentage points under 128, 64, and 32 tokens, respectively.
Such results confirm that preserving complementary visual token roles is particularly effective under limited token budgets.

\begin{table}[tb]
    \centering
    \footnotesize
    \setlength{\tabcolsep}{1.0mm}
    \resizebox{1\linewidth}{!}{
        \begin{tabular}{@{}l|ccccc|cc@{}}
        \toprule
        Method & VQAv2 & GQA & VizWiz & SQA & VQA$^{\mathrm{Text}}$ & Avg. & Rel. (\%)\\
        \midrule
        \multicolumn{8}{c}{Qwen-VL-7B (128/256, $\downarrow$ 50\%)} \\
        \cmidrule{1-8}
        Qwen-VL & 78.8 & 59.3 & 35.2 & 67.1 & 63.8 & 60.8 & 100.0\% \\
        FastV & 73.3 & 55.2 & 35.2 & 67.3 & 59.7 & 58.1 & 96.0\% \\
        DivPrune & 77.8 & 57.8 & 33.0 & 67.7 & 56.6 & 58.6 & 95.9\% \\
        VisPruner & 77.1 & 57.6 & 30.9 & 67.7 & 55.4 & 57.7 & 94.1\% \\
        Ours & 78.1 & 58.4 & 33.2 & 68.4 & 61.2 & \textbf{59.9} & \textbf{98.0\%} \\
        \midrule
        \multicolumn{8}{c}{InternVL-13B (144/576, $\downarrow$ 75\%)} \\
        \cmidrule{1-8}
        InternVL & 79.3 & 62.9 & 52.6 & 66.3 & 57.0 & 63.6 & 100.0\% \\
        FastV & 74.3 & 58.4 & 51.4 & 66.6 & 55.6 & 61.3 & 96.5\% \\
        DivPrune & 76.4 & 60.6 & 54.5 & 67.2 & 54.4 & 62.6 & 98.6\% \\
        VisPruner & 76.2 & 60.4 & 53.2 & 67.2 & 54.5 & 62.3 & 98.0\% \\
        Ours & 77.1 & 60.7 & 53.6 & 67.4 & 55.4 & \textbf{62.8} & \textbf{98.9\%} \\
        \midrule
        \multicolumn{8}{c}{CogVLM-7B (123/1225, $\downarrow$ 90\%)} \\
        \cmidrule{1-8}
        CogVLM & 80.0 & 59.4 & 38.9 & 70.5 & 69.2 & 63.8 & 100.0\% \\
        FastV & 59.7 & 42.2 & 25.4 & 66.6 & 26.4 & 44.1 & 68.6\% \\
        DivPrune & 76.4 & 54.8 & 35.4 & 70.1 & 59.7 & 59.3 & 92.7\% \\
        VisPruner & 72.0 & 52.2 & 32.2 & 68.9 & 52.9 & 55.6 & 86.8\% \\
        Ours & 75.8 & 52.5 & 35.8 & 70.4 & 63.2 & \textbf{59.5} & \textbf{93.1\%} \\
        \bottomrule
        \end{tabular}
    }
    \caption{
        Performance across LVLM architectures. Avg. denotes the normalized average score across benchmarks, and Rel. denotes the relative performance retention ratio compared with the dense model.
    }
    \label{tab:multiarch}
\end{table}

\subsection{BtsPruner in High-Resolution Settings}

High-resolution LVLMs generate substantially longer visual-token sequences, making token pruning particularly important.
We apply BtsPruner to LLaVA-NeXT-7B, which processes 2880 visual tokens.
As stated in Table~\ref{tab:highres}, BtsPruner achieves the highest performance retention across all pruning ratios.
When retaining 640 tokens, BtsPruner preserves 99.1\% of the dense model performance.
Even with only 160 retained tokens, it maintains 91.3\% performance retention, demonstrating that preserving complementary token roles remains effective under high-resolution token compression.

\subsection{BtsPruner across Diverse LVLM Architectures}

To evaluate the general applicability of BtsPruner, we further apply it to LVLMs with different projector designs, including Qwen-VL~\cite{bai2023qwenvl}, InternVL~\cite{10656429}, and CogVLM~\cite{wang2025cogvlm} as summarized in Table~\ref{tab:multiarch}. These models adopt different visual-language alignment strategies, providing a comprehensive evaluation beyond the LLaVA series.

For Qwen-VL, which employs a query-based projector, we introduce a lightweight adaptation. Specifically, after pruning visual tokens with BtsPruner, we select query tokens according to their attention responses over the retained visual tokens. For InternVL and CogVLM, which use MLP-based projectors, BtsPruner is directly applied after the vision encoder.
Across various architectures, BtsPruner achieves the highest Avg. and Rel. among comparison methods, demonstrating that the proposed role-aware token preservation strategy generalizes across different LVLM architectures.

\begin{table}[t]
    \small
    \centering\footnotesize\setlength{\tabcolsep}{1.0mm}
    \begin{tabular}{@{}c|c|c|c|c@{}}
    \toprule
    \textit{Global Rep.} & \textit{Local Detail} & TRIM & Diverse & Avg. \\
    \midrule
    Random & Random & $\times$ & $\times$ & 48.5 \\ \hline
    \multirow{5}{*}{Our} & Random & $\times$ & $\times$ & 54.7 \\ \cline{2-5}
    & \multirow{4}{*}{Our} & $\times$ & $\times$ & 55.3 \\
    & & $\surd$ & $\times$ & 55.6 \\
    & & $\times$ & $\surd$ & 56.1 \\
    & & $\surd$ & $\surd$ & \textbf{56.3} \\
    \bottomrule
    \end{tabular}
    \caption{
        Ablation study of BtsPruner components on TextVQA. \textit{Global Rep.} and \textit{Local Detail} denote global representative discovery and local detail recovery, respectively.
    }
    \label{tab:table4_deep_shallow}
\end{table}

\begin{table}[tb]
    \centering
    \footnotesize
    \setlength{\tabcolsep}{1.5mm}
    \begin{tabular}{@{}c | c c c@{}}
        \toprule
        \# Token & $r$ & VisPruner & BtsPruner \\
        \midrule
        & 0.25 & 56.7 & \textbf{57.7} \\
        128 & 0.50 & 57.2 & \underline{57.7} \\
        & 0.75 & 56.8 & 57.3 \\
        \midrule
        & 0.25 & 55.7 & \textbf{56.3} \\
        64 & 0.50 & 55.8 & \underline{56.3} \\
        & 0.75 & 56.0 & 56.2 \\
        \midrule
        & 0.25 & 53.2 & 53.5 \\
        32 & 0.50 & 53.5 & \textbf{54.5} \\
        & 0.75 & 53.3 & \underline{54.0} \\
        \bottomrule
    \end{tabular}
    \caption{Influence of the global-representative token ratio $r$ on TextVQA performance.}
    \label{tab:table5_retain_tokens}
\end{table}

\begin{table}[t]
    \centering
    \footnotesize
    \setlength{\tabcolsep}{1.0mm}
    \resizebox{1\linewidth}{!}{
        \begin{tabular}{@{}l|c|c|c|c|c@{}}
        \toprule
        \multicolumn{1}{l|}{\parbox[c][3.5ex][t]{10mm}{\centering Method}} & \multicolumn{1}{c|}{\parbox[c][3.5ex][t]{10mm}{\centering \# Token}} & \shortstack{FLOPs \\
        ↓ (T)} &
        \shortstack{Cache \\
        ↓ (MB)} &
        \shortstack{GPU Mem. \\
        ↓ (GB)} &
        \shortstack{CUDA Lat. \\
        ↓ (ms)} \\
        \midrule
        LLaVA-NeXT & 2880 & 40.8 & 1471 & 17.8 & 821 \\
        \midrule
        FastV & 640 & 13.2 & \underline{370} & 17.6 & 332 \\
        DivPrune & 640 & \underline{11.4} & \textbf{351} & \textbf{15.2} & 282 \\
        VisPruner & 640 & \textbf{11.1} & \textbf{351} & \underline{15.5} & \textbf{264} \\
        Ours & 640 & \textbf{11.1} & \textbf{351} & \underline{15.5} & \underline{270} \\
        \midrule
        FastV & 160 & 7.4 & \underline{125} & 17.6 & 207 \\
        DivPrune & 160 & \underline{4.9} & \textbf{111} & \textbf{14.7} & 157 \\
        VisPruner & 160 & \textbf{4.8} & \textbf{111} & \underline{15.4} & \underline{145} \\
        Ours & 160 & \textbf{4.8} & \textbf{111} & \underline{15.4} & \textbf{138} \\
        \bottomrule
        \end{tabular}
    }
    \caption{
        Efficiency comparison on LLaVA-NeXT-7B using the POPE benchmark. Latency is measured as CUDA inference latency on a single RTX 3090 GPU.
    }
    \label{tab6:efficiency}
\end{table}

\subsection{Ablation Study}

We conduct ablation experiments to evaluate the contribution of each component in BtsPruner and analyze the influence of the global-representative token ratio $r$. Table~\ref{tab:table4_deep_shallow} presents the component ablation results on TextVQA using LLaVA-1.5-7B, where $r=0.5$ and 64 visual tokens are retained. Starting from random selection, introducing global representative discovery improves the Avg. from 48.5 to 54.7, demonstrating the importance of preserving globally referenced tokens. Adding local detail recovery further improves the performance to 55.3, indicating that local details provide complementary visual evidence. TRIM and Diverse bring additional improvements, and their combination achieves the best performance of 56.3.
These results verify the effectiveness of each component in BtsPruner.

We further investigate the influence of the role allocation ratio $r$. Following the same setting, we evaluate $r\in\{0.25,0.5,0.75\}$ under different token budgets. As stated in Table~\ref{tab:table5_retain_tokens}, BtsPruner remains stable across different values of $r$ and consistently outperforms VisPruner. This indicates that preserving both global representatives and local details is more effective than relying on a single token selection strategy.

\subsection{Efficiency Analysis}

To evaluate the computational efficiency of BtsPruner, we compare it with FastV, DivPrune, and VisPruner on LLaVA-NeXT-7B in terms of FLOPs, cache usage, GPU memory, and CUDA latency. The experiments are conducted on a single NVIDIA GeForce RTX 3090 GPU with the POPE benchmark.
As stated in Table~\ref{tab6:efficiency}, when retaining only 160 visual tokens, BtsPruner reduces CUDA latency from 821 ms to 138 ms, achieving a 5.95$\times$ speedup. Compared with existing pruning methods, BtsPruner introduces negligible additional overhead while maintaining better role preservation and performance retention under aggressive token reduction.

\section{Conclusion}

We propose BtsPruner, a training-free role-aware visual token pruning framework for LVLMs. By analyzing the evolution of visual token roles during encoding, we observe that visual tokens gradually transition from local detail carriers to globally referenced representatives. Based on this observation, BtsPruner identifies the role transition point, discovers global representative tokens, and recursively retrieves complementary local detail tokens under a fixed token budget. Extensive experiments across multiple LVLM architectures, multimodal benchmarks, and pruning ratios demonstrate that BtsPruner effectively preserves complementary visual token roles under aggressive compression. Without training, BtsPruner retains over 91\% of the dense model performance while keeping only 5.6\% of the visual tokens, and achieves a 5.95$\times$ CUDA-latency speedup in high-resolution settings.


\bibliography{ref}


\appendix
\onecolumn
\part*{\centering Supplementary Material \\ for \\ BtsPruner: Role-Aware Visual Token Pruning for Large Vision-Language Models}

This appendix is organized as follows.
\begin{itemize}
\item In Section A, additional visualization results are presented.
\end{itemize}

\section{Visualization}

\subsection{Visualization comparison of visual token pruning strategies}

\begin{figure}[!b] 
	\centering
	\includegraphics[width=\columnwidth]{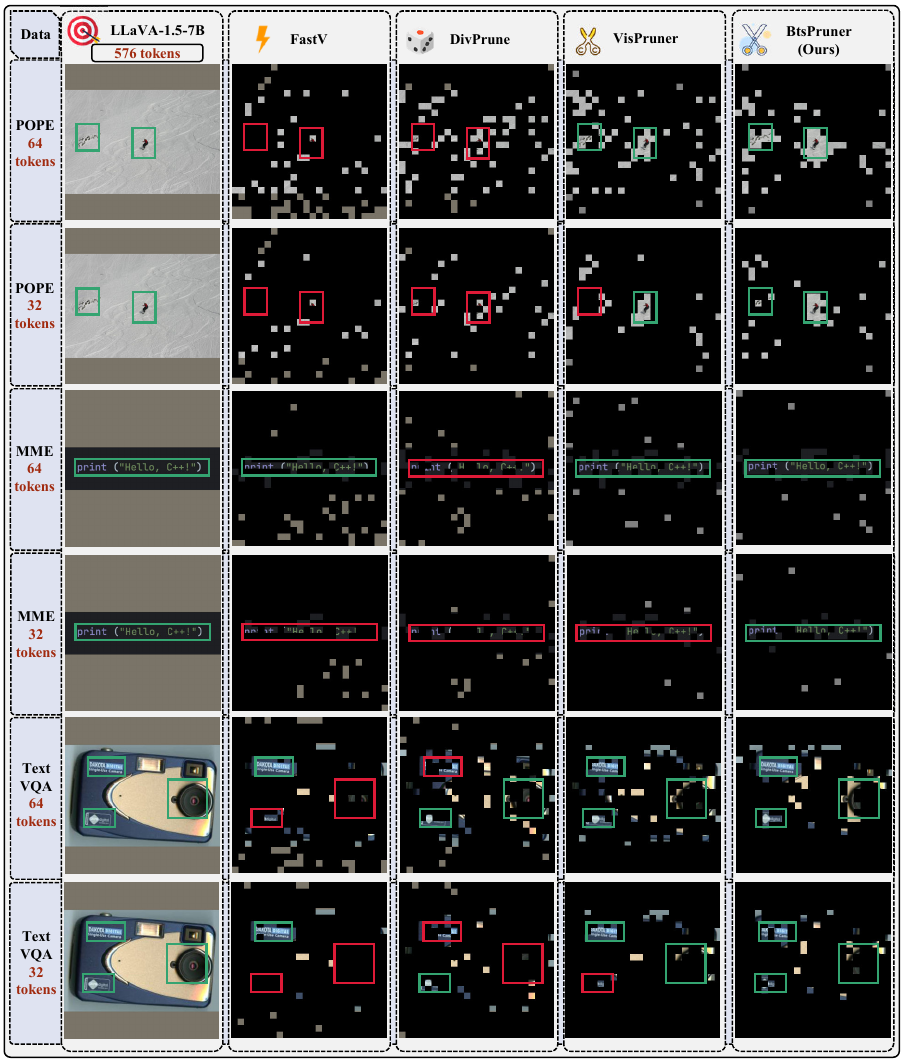}
	\caption{Qualitative comparison of visual token pruning strategies.}
	\label{visualization}
\end{figure}

To further validate the effectiveness of BtsPruner for visual token pruning, we present more comprehensive visualizations of the baseline networks \cite{liu2024llava1.5} under different pruning methods. As shown in Fig.~\ref{visualization}, existing pruning strategies \cite{chen2024fastv, 11095009, 11444044} evaluate all visual tokens with a unified criterion, making them insensitive to the distinct roles played by different tokens. Consequently, given the same token budget, they may either miss globally representative tokens or discard tokens that encode rich local fine-grained details. In contrast, BtsPruner adopts a role-aware selection mechanism that preserves complementary visual information in a more comprehensive manner.

\subsection{Supplementary Visualizations of Attention Patterns}

\begin{figure}[!t]
	\centering
	\includegraphics[width=\columnwidth]{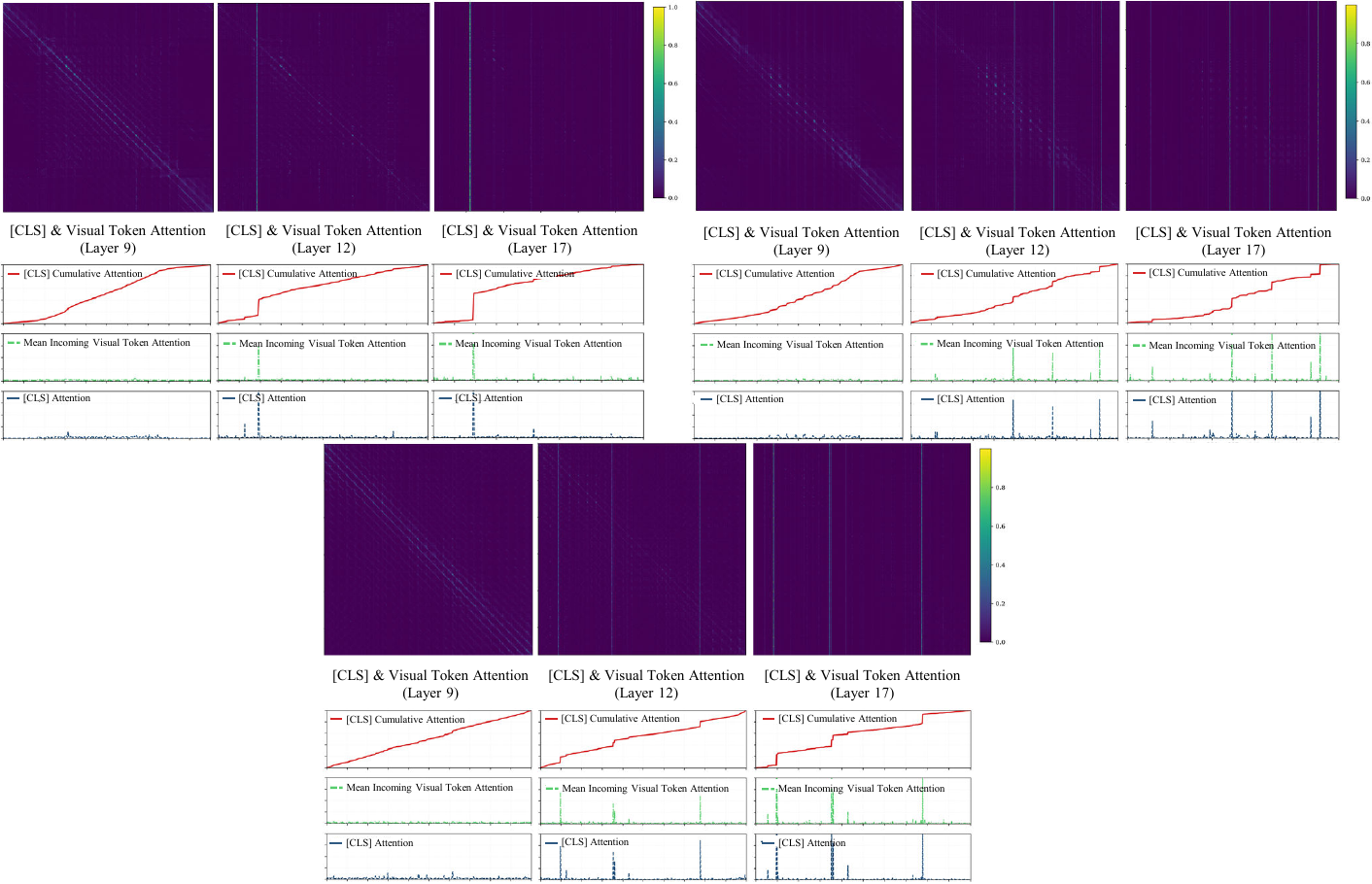}
	\caption{Visualization of attention evolution during the encoding process across different image samples.}
	\label{evolution}
\end{figure}

In this section, we present attention visualizations during the encoding process across different image samples. As shown in Fig.~\ref{evolution}, the results exhibit a broadly consistent attention pattern, progressing from localized interactions to shared representations.

\end{document}